# DeepDA: LSTM-based Deep Data Association Network for Multi-Targets Tracking in Clutter

Huajun Liu
*School of Computer Science and Engineering, Nanjing University of Science and Technology Robotics Institute, Carnegie Mellon University*
liuhj@njust.edu.cn

Hui Zhang
*School of Computer Science and Engineering, Nanjing University of Science and Technology*
huizh@njust.edu.cn

Christoph Mertz
*Robotics Institute, Carnegie Mellon University*
cmertz@andrew.cmu.edu

*Abstract*—The Long Short-Term Memory (LSTM) neural network based data association algorithm named as DeepDA for multi-target tracking in clutter is proposed to deal with the NP-hard combinatorial optimization problem in this paper. Different from the classical data association methods involving complex models and accurate prior knowledge on clutter density, filter covariance or associated gating etc, data-driven deep learning methods have been extensively researched for this topic. Firstly, data association mathematical problem for multi-target tracking on unknown target number, missed detection and clutter, which is beyond one-to-one mapping between observations and targets is redefined formally. Subsequently, an LSTM network is designed to learn the measurement-to-track association probability from radar noisy measurements and existing tracks. Moreover, an LSTM-based data-driven deep neural network after a supervised training through the BPTT and RMSprop optimization method can get the association probability directly. Experimental results on simulated data show a significant performance on association ratio, target ID switching and time-consuming for tracking multiple targets even they are crossing each other in the complicated clutter environment.

## I. INTRODUCTION

Multi-Targets Tracking (MTT) consists of automatically excluding clutter and generating tracks from unlabeled data sequences. It has numerous applications in radar target detection, tracking, and recognition [1], and track-level sensor fusion [2]. It is also usually used as a part of various LiDAR applications (moving object tracking) [3], autonomous vehicles [4], advanced driver assistance system [5], and simultaneous localization and mapping [6] (using optical, *e.g.*, camera or LiDAR, or radar sensors) for robotics. Data association, one of the most challenging components of MTT, examines the relations between sensor measurements and the targets. Specifically, it main focus on how to determine which observations should be assigned to which target. A measurement may be caused by one or multiple targets, or by the environment (false alarm/clutter), and sometimes a real target may be missed-detected because of the insufficient detection probability.

The goal of data association is to identify a correspondence between sensor measurements and targets. New measurements can be generated by previously undetected targets or pre-existing tracks, so the measurement-target correspondence must consider a track initialization. Likewise, the measurements that stem from clutter within the surveillance region must be identified to avoid false alarms.

A common way to formulate these data association tasks is as an assignment problem [7]. The simplest version is the 2D assignment problem, also termed bipartite matching or 0-1 integer programming (IP), which seeks to match $m$ tracks to $n$ measurements. This combinatorial optimization problem ensures that each track is assigned to exactly one measurement, but measurements are not allowed to be assigned to clutter, which would cause false alarms or to be assigned to a dummy track which would cause a missed detection.

Fig. 1 shows a typical scenario describing the measurements by a radar where there are false alarms and miss detections. In the Fig. 1, within the dwell volume there are three measurements generated by targets, three false alarms caused by the environment, one missed detection which can't be detected by the radar, and two closely spaced targets that are not resolved by a resolution cell. These false alarms, missed detection or unresolved targets belong to a type of uncertain information, and this type of information uncertainty caused by sensors is known as measurement origin uncertainty [8].

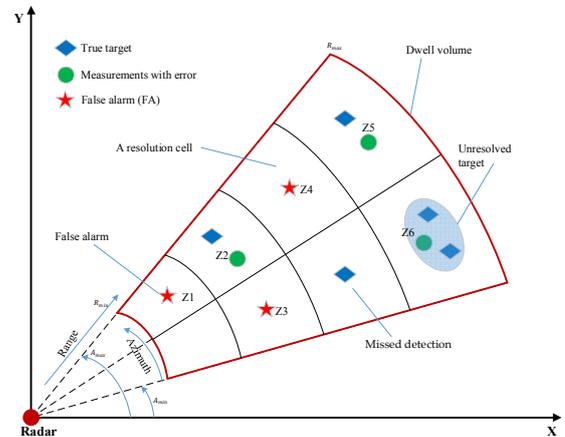

Fig. 1. A typical radar measurement scenario [8]

An early contribution in this topic can date from the 1970s, when C. L. Morefield applied 0-1 integer programming to the data association problem (DAP) [9]. Subsequently, the DAP is further extended to a multidimensional assignment problem (MDAP) [9, 10, 11], and the notable Lagrangian relaxation is suggested to solve the problem as well.

Generally, two types of approaches are investigated to the DAP in multi-target tracking. One is called the target-oriented approach, and another is called the measurement-oriented approach. In the former, each measurement is assumed to have originated from either a known target or clutter, and classical methods such as probabilistic data association (PDA) [1, 10, 12] and joint probabilistic data association (JPDA) [1, 13, 14, 15] belong to this type. In the latter, each measurement is assumed to have originated from either a pre-existing target, a new target, or clutter [10]. Moreover, multi-scan methods such as multi-hypotheses tracking [16, 17], smoothing multi-scan [18] are preferable where the objects of interest are either closely spaced, or with amount of clutter, or with extremely low detection probability. However, delaying the association decisions with multiple dwells or multiple hypothesis would negatively affect the real-time capabilities of the tracker.

Measurement-to-track association is an instance of NP-hard combinatorial optimization problem known as the MDAP [7, 19]. Using binary decision variables to represent an assignment of a measurement to a track, then it can be simplified into a 0-1 integer program, and the goal is to find the optimal assignment of measurements to tracks so that the total assignment cost is minimized. To capture the fact that some sensor measurements will either be false alarms or missed detections, some gating techniques, such as elliptical gates, rectangular gates etc., can be used to determine, with some degree of confidence, whether any of the new measurements might have originated from a track. Even though there are $mn!$ possible assignments, many polynomial-time algorithms exist for finding the globally optimal assignment matrix.

Most famous is the Hungarian algorithm (HA) [7, 20] whose computational complexity is $O(N^3)$, and similar algorithms such as Kuhn-Munkres and Auction [7, 20] are also fast and easy to integrate into real-time MTT systems. It is assumed that there are N measurements at time t, the Hungarian algorithm can get the mapping between target and measurement in $O(N^3)$ time. Hungarian algorithm solves the problem of weighted bipartite graph matching. One side of the node of the bipartite graph is a set of measurements, and the other side of the set is a set of target prediction states, the weight of the arc between the node representing the i-th target state and the j-th measured value is a log-likelihood function, and the Hungarian algorithm minimizes the estimated allocation cost of the target by maximizing the sum of the log likelihood function. Since all measurements are compared to all prediction states, this method can also be called a global nearest neighbor (GNN) method. But they do not yield good results in a complex scenario, especially in crowded scenes with clutter or occlu-sion.

Efficient approximations are also considered for data association. Approaches like probabilistic data association (PDA) [12] or joint probability data association (JPDA) [13] are also used to data association in clutter. The main idea of PDA is that all the measurements that fall into the associated gates may come from the target with different probabilities, and use these probabilities to calculate the weighted summation of each effective measure. The equivalent echo is used to update the target state. Based on the PDA algorithm, Bar-Shalom *et al.* [1] proposed an algorithm named as JPDA that introduces the concept of "cluster", which becomes one of the classic algorithms for data association in complex multi-target tracking. The JPDA algorithm needs to calculate the probability of all possible events, and the corresponding validation matrix split and calculation would have a combined explosion problem [20]. Meanwhile, solving the optimal data association is an NP-hard problem, and at the same time, with the number of targets and measurement increasing, the computational burden of the JPDA algorithm is larger and larger.

Neural networks have a rich history of being used to solve combinatorial optimization problems. One of the most influential papers in this line of research, by Hopfield and Tank [21] uses Hopfield nets to approximately solve instances of the Traveling Salesman Problem (TSP). Wang *et al.* [22] considered a Recurrent Neural Network (RNN) to solve the assignment problem, however, the number of iterations to achieve an optimal solution increased with the problem scale. Researchers include Mengyuan Lee [23] and Saïd Medjkouh [24] *et al* proposed different deep neural networks for assignment problem resolver, but their assignment is subjected to the 1-1 constraint. Anton Milan *et al.* proposed a RNN for TSP resolver [25], and many other RNNs [26-32] are also proposed for multi-target tracking. But most of these RNNs [26, 28, 30] limit that it can only be used in a clear and reliable environment without clutter or missed-detections. And one of the key challenges that these supervised learning approaches face in this domain is obtaining labeled ground-truth samples, since generating optimal solutions to NP-hard combinatorial optimization problems can be time-consuming or even impossible. And Pol Rosello [33] *et al* uses reinforcement learning for multi-target tracking to automatically manage the track's birth-to-death, but its data association uses Hungarian Algorithm. An alternative to this is the approach in [34] tries to use a collaborative reinforcement learning for multi-target tracking in a visual scene. The main difficulties here are deciding how to represent the data for efficient learning and enforcing the original constraints of the problem during training. In other words, given noisy measurements of the environment, it is also very difficult for a deep learning system to directly output the filtered tracks, combining the association problem with state estimation. Research on applying deep learning to the DAP in multi-target tracking is still in its infancy.

In this paper, we construct a data-driven LSTM-based data association network for multiple targets tracking in clutter, which calculates the association probability of measurements to tracks compatible with 1-0 and 0-1 constraints to satisfy the requirement of data association in clutter and missed-detections.

Our contributions can be summarized as follows. First, we extend an LSTM network of data association problems with 1-1 constraint to be compatible with 1-0 and 0-1 constraints to satisfy the requirement of data association in clutter and missed-detections. Second, it is shown that our data association framework without any prior knowledge is a supervised sequence-to-sequence learnable scheme which can convert the integer programming problem to an association probability estimation problem. In addition, to the best of our knowledge, this is one of first papers addressing the data association for multi-target tracking in dense clutter and missed-detections with a deep neural network, the results on clutter scenario show remarkable potential.

## II. PROBLEM FORMATION

A radar or LiDAR is assumed to receive measurements periodically. Suppose that a radar system starts scanning a region, then the set of measurements is recoded within the interval $[t_0, t_1]$, where $t_0$ is the time when scan starts, and $t_1$ is the time when scan ends. Seen as in Fig. 2, it is a particular shot of a radar or Lidar's data association between a measurement set and a target set.

The measurement set $Z(k)$ [9] in scan k is defined as follows:

$$Z(k) = \{z_{i_k}^k\}_{i_k=1}^{M_k}, \text{ for all } k = 1, \ldots, N \quad (1)$$

Here, $N$ is the number of scans, $M_k$ is the total number of measurements received in scan $k$ and $z_{i_k}^k$ is the $i$-th measurement vector in scan $k$, containing range, azimuth, and even elevation data with errors.

In order to model false alarms and missed detections, the concept of a dummy report [9] is introduced. By adding a dummy measurement, the Equ. (1) is redefined where indexed $i_k = 0$ represents the false alarms. Then the measurement sets $Z(k)$ is redefined formally:

$$Z(k) = \{z_{i_k}^k\}_{i_k=0}^{M_k}, \text{ for all } k = 1, \ldots, N \quad (2)$$

The target set $\Gamma(k)$ in scan $k$ is defined as follows:

$$\Gamma(k) = \{\tau_{j_k}^k\}_{j_k=1}^{N_k}, \text{ for all } k = 1, \ldots, N \quad (3)$$

Here, $\tau_{j_k}^k$ is the $j$-th exist target state prediction to scan $k$, $N_k$ is the total number of target has been initiated and maintained until scan $k$.

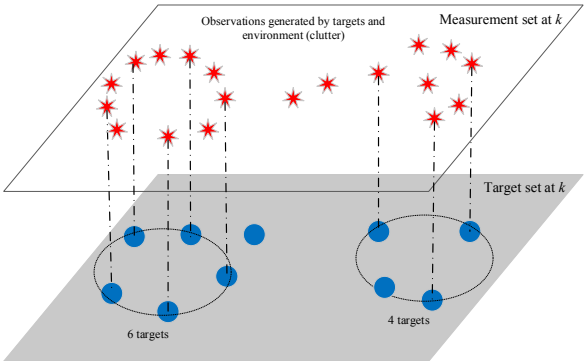

Fig. 2. Data association between measurement set and target set

We define the pairwise-distance matrix $C \in \mathbb{R}^{N \times M}$ to calculate the distance of all measurements and predicted state of all targets. And for any measurement $z_{i_k}^k \in Z(k)$, and any target $j$, $\tau_{j_k}^k \in \Gamma(k)$, the Euclidean distance between the measurement $i$ and the predicted state of target $j$ is $C_{ij}^k \in C$, which is defined as

$$C_{ij}^k = \|z_{i_k}^k - \tau_{j_k}^k\|_2 \quad (4)$$

For each $z_{i_k}^k \in Z(k)$, a 0-1 decision variable $x_{i_k j_k}^k$ is introduced, which is defined as follows,

$$x_{i_k j_k}^k = \begin{cases} 1, & \text{if } z_{i_k}^k \text{ is assigned to } \tau_{j_k}^k \text{ at time } k \\ 0, & \text{otherwise} \end{cases} \quad (5)$$

The complete DAP model of radar tracking in clutter with $N$ scans can now be defined as follows:

$$\text{Minimize } \sum_{k=1}^{N} \sum_{i_k=0}^{M_k} \sum_{j_k=1}^{N_k} C_{ij}^k x_{i_k j_k}^k \quad (6)$$

Subject to

1) $\sum_{j_k=1}^{N_k} x_{i_k j_k}^k \leq 1$, for all $k = 1, \ldots, N$
   $x_{ij} \in \{0,1\}$

2) $\sum_{i_k=0}^{M_k} x_{i_k j_k}^k \leq 1$, for all $k = 1, \ldots, N$
   $x_{ij} \in \{0,1\}$

The constraint 1) means each measurement can only be assigned to at most one target, in other words, some measurements originated from clutter would not be associated with any target; and the constraint 2) means each target could only be associated with at most one measurement, in other words, some targets would not be associated with any measurements because of missed-detections.

That means the data association for multi-target tracking in clutter is a DAP with constraints besides 1-1 case (one measurement originated from one target), also with 0-1 case (missed-detections) and 1-0 case (clutter or false alarm). It is easy to see that the DAP problem can be reformulated exactly as a complex combinatorial optimization problem [35, 36, 37].

### III. LSTM-BASED DEEP DATA ASSOCIATION NETWORK

#### A. LSTM-based Sequence-to-Sequence Data Association Network

A data driven LSTM architecture that can learn to solve combinatorial optimization problem entirely from training samples for data association was proposed in this paper. This is of great significance for many reasons. First, data association in clutter is generally a highly complex, discrete combinatorial optimization problem which is a NP-hard problem. Second, most solutions in the output space are merely permutations of each other w.r.t. the input features. Finally, real data association should meet the 1-1 constraint to prevent the same measurement to be assigned to multiple targets, meet the 1-0 constraint including the situation of missed-detection, and meet the 0-1 constraint including environment clutter or false alarms also. We believe that the LSTM can be learned effectively because of its non-linear transformations and strong memory component. To support this point, we extend the LSTM-based data association to resolve the linear assignment problem.

After each scan dwell of sensor measurement, the input of LSTM network can be preprocessed by a pair-wise distance matrix generated from all measurements and state prediction of each target. Then measure-to-track association probability is calculated by the LSTM networks. Each target is updated by the associated measurement to maintain a continuous and complete track. The overview of proposed deep data association framework is seen as Fig. 3.

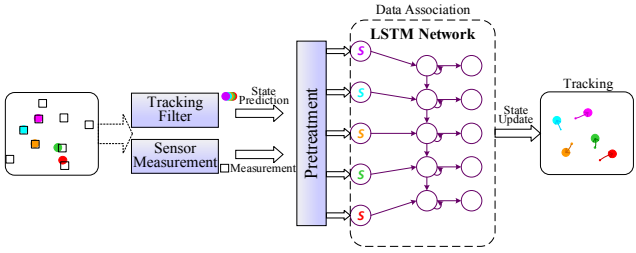

Fig. 3. Overview of our proposed data association framework

The main idea is to exploit the LSTM's temporal step-by-step functionality to predict the assignment for each measurement to one target at a time. The input at each step $i$, next to the hidden state $h_i$ and the cell state $c_i$, are the entire feature vector. The input of the network is a pairwise-distance matrix which is an Euclidean distance matrix between the predicted state of target $i$ and measurement $j$. Note that it is straight-forward to extend the feature vector to incorporate appearance or any other similarity information. The output that we are interested in is then a vector of probabilities $A^i$ for one target and all available measurements, obtained by applying a softmax layer with normalization to the predicted values. Here, $A^i$ denotes the $i^{th}$ row of A.

The overview of our approach is illustrated in Figure 4. The main idea is to design a data association model based on LSTM. This deep neural network model can learn the method of predicting the associated probability matrix completely from the data. The LSTM network is designed for sequence-to-sequence association and prediction [25], each time only predicting the association probability of $N$ target and $M$ measurements.

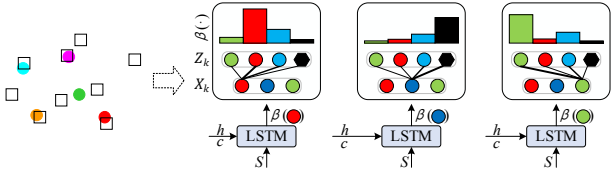

Fig. 4. Diagram of data association model based on LSTMs

Predicting the probability of association between each target and all measurements in one time is shown in Fig. 4. In each prediction step, the network will output a vector of probability distribution, which is the association probability of a target with all measurement at time $k$, i.e., $\beta_{i\cdot} \in \mathbb{R}^{1\cdot(M+1)}$, where an extra column indicates that the target has missing measurements.

The input of the network is $S_i \in \mathbb{R}^{1\times(D*M)}$ which passes through a fully connected layer to the hidden state $h_i$. And the output is the association probability $\beta_{i\cdot} \in \mathbb{R}^{1\cdot(M+1)}$ to the $i$-th target which is obtained through a fully connected layer transformation and a subsequent Sigmoid transformation. The specific structure of the network, which references the design of [26] is shown in Fig. 5.

The input vector $S_i$ is reshaped by a distance matrix which is the distance matric of all measurements to all targets. The reshape on $S_i$ is calculated as follows:

$$S_i = \text{reshape}\{\text{repmat}((Hx_k^i)^T, M, 1) - Z_k, 1, D*M\} \quad (7)$$

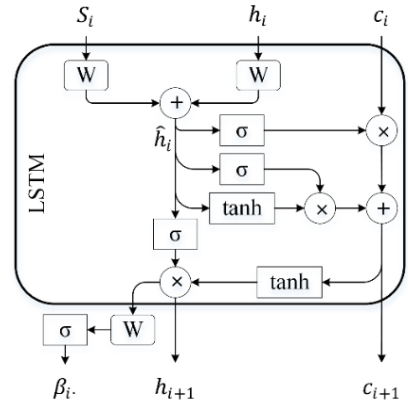

Fig. 5. Structure of LSTM-based data association unit

where $(Hx_k^i)^T \in \mathbb{R}^{1\times D}$ is the prediction of the measurement of the target $i$, and $\text{repmat}((Hx_k^i)^T, M, 1)$ indicates to repeat $(Hx_k^i)^T$ $M\times 1$ times to form a $M\times D$ dimensional matrix.

The algorithm uses the Mean Square Error (MSE) function as the loss function to minimize the MSE between the predicted association probability and the real target-measurement association probability. The loss function is defined as

$$\mathcal{L}(\beta_{i\cdot}, \tilde{\beta}_{i\cdot}) = \sum\|\beta_{i\cdot} - \tilde{\beta}_{i\cdot}\|^2 \quad (8)$$

Where $\tilde{\beta}_{i\cdot}$ represents the ground truth of the association probability of the $i$-th target with all measurements.

At each prediction step, the network outputs a probability distribution that represents the probability of association of a target with all observation sets $Z_k$ at time k, and the probability that no measurement (hexagon in the Fig. 4) is associated with the target.

### B. Training Algorithm and Parameters

The DeepDA network is trained by a supervised algorithm in a multi-target tracking in clutter environments. The Back-propagation Through Time (BPTT) algorithm [38] combined with the RMSprop optimization algorithm [39] is used to train the network.

The training parameters are divided into two parts: one is for the network parameter, which mainly determines the size of the input to output size in each network layer in the whole model; the other part is the training parameter, such as the learning rate and decay rate, which determines the training error and efficiency of the network.

The parameters of the network are trained by a supervised training algorithm, which can learn from data samples of the target ground truth and measurements with clutter and missed-detections, and the output is a vector of the association probability of the measurements to target. The overall framework of the network model is shown in Fig. 6, where W is the fully connected layer. And $X_t, t = 1,2,\cdots,T$ represents all input states at a time, $P_t, t = 1,2,\cdots,T$ indicates the output status at a certain time.

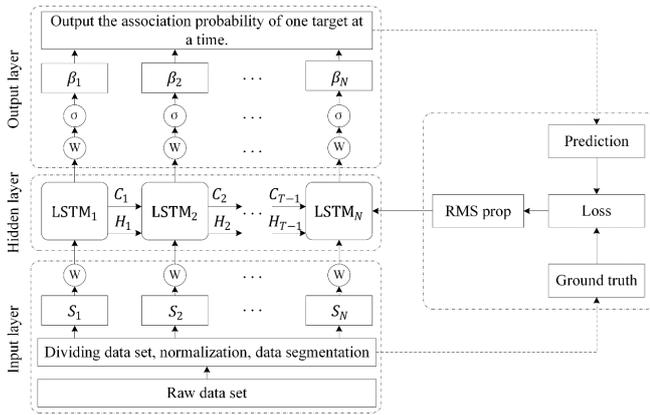

Fig. 6. LSTM-based DeepDA training framework

The dataflow of the DeepDA algorithm's training is as follows:

**Algorithm: Training algorithm of DeepDA Network**

Input: the ground truth of a target at a certain moment, and its corresponding next-time observation;
Output: the association probability of this target with current measurements.
Step 1: Set the data parameters, select and normalize the data by min-max method according to the min-max conversion function $X^* = (X - \min)/(\max - \min)$;
Step 2: Set the network structure parameters, determine the size of the network structure, and initialize the node status of each layer, generally all set to 0;
Step 3: Set training parameters and start the network training, e=1: epoch:
  (a) Randomly select data for one batch;
  (b) Forward computation: Calculate the output value of each LSTM in turn as the number of time windows increases;
  (c) Calculate the error according to the loss function;
  (d) Backpropagation: Backpropagation of the error term of each LSTM, with propagation in two directions: one is backpropagation in the spatial direction, i.e., to the upper layers; one is backpropagation along decrease direction with the batch size;
  (e) Optimize network parameters using the RMSprop method;
  (f) e = e+1;
Step 4: End the loop.

## IV. EXPERIMENTAL RESULTS AND ANALYSIS

### A. Experiment Setup and Data Collection

In order to completely evaluate the data association performance of our proposal with conditional algorithms, we setup a simulated experimental scenario. Similar in [25], it is assumed that there are 5 targets placed in a 2-D space, and they are placed in the range of 4m to 25m along the x axis, and in the range of 10m to 20m along the y axis. And their trajectories would be crossing each other at a certain time. Through the simulation experiments, the effectiveness and accuracy of the proposed algorithm are proved.

In the scenario, the 5 targets will move along a line in different directions separately within the specified range. The measurement sampling interval is 1 plot per second, the total sampling number is 20, and 5 targets will cross each other within the 9th to 12th sampling time. The measurements are generated by the sampling plots multiplying by a detection probability $P_d$, at the same time the clutter which is generated by a Poisson distribution with a clutter density $E_\lambda$ within the surveillance area is superimposed to the tracks. And our clutter density $E_\lambda$ is calculated as the number of clutter points per unit area.

The initial state of these 5 targets is set according to the TABLE I.

TABLE I.     INITIAL PARAMETERS OF 5 TARGETS IN THE SIMULATION SCENARIO

|  | Target 1 | Target 2 | Target 3 | Target 4 | Target 5 |
|---|---|---|---|---|---|
| x/m | 5 | 5 | 5 | 5 | 5 |
| vx/m² | 1 | 1 | 1 | 1 | 1 |
| y/m | 11 | 13 | 15 | 17 | 19 |
| vy/m² | 0.4 | 0.2 | 0 | -0.2 | -0.4 |

The measurements are obtained by adding Gaussian white noise with variance size $\sigma_x = 0.3162m$ and $\sigma_y = 0.3162m$ on x and y axis. The target motion trajectory of the simulated scene and the corresponding measurement are shown in Fig. 7. Different color in the figure corresponds to a specific target.

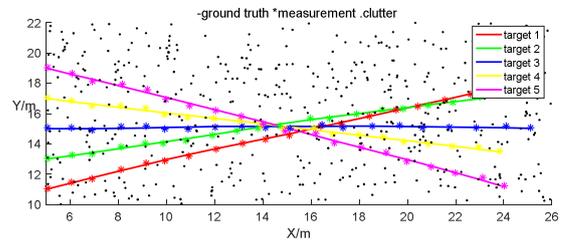

Fig. 7. Simulation of real trajectories and measurements

The training data is generated through above simulation environment. Specifically, in a sequence at time k, the training data will include: the input state of the model, all the measurements $Z_k$ and its predicted state $X_k$ at time k, and the real associated probability $\beta^k$, if the measurement is originated from the target, $\beta^k = 1$, else $\beta^k = 0$.

### B. Performance Metric

There are three evaluation indices adopted in the experiments to evaluate the different data association algorithms, which are the Optimal Sub-Patten Assignment (OSPA) distance, the Switch Times of Target ID (STTI), and time consuming of the algorithm.

OSPA is an important metric to evaluate the overall performance of the multi-target tracking system [40]. It provides a measure to quantitatively evaluate the difference between the real trajectory and the estimated trajectory in multi-target tracking system. It is better than the conditionally used Root Mean Square Error (RMSE) for different tracking algorithms in precision and sensitivity. And computation of RMSE requires a determined association relationship between different target trajectories first, but this is exactly what we lack.

It is assumed that the set of targets with real state is $X = \{x_1, x_2, \cdots x_m\}$, the set of estimated targets is $\hat{X} = \{\hat{x}_1, \hat{x}_2, \cdots \hat{x}_n\}$, where $m$ and $n$ respectively indicates the ground truth of targets number and the estimated targets number; $x \in X, \hat{x} \in \hat{X}$ are elements in the two sets which are the real target state vector and the estimated target state vector at a certain time respectively.

The OSPA distance calculation is defined as follows [40].

$$OSPA_{p,c}(X,\hat{X}) =$$

$$\left[\frac{1}{n}\left(\min_{\pi\in\Pi_n}\sum_{i=1}^{m}\left(d_c(x_i,\hat{x}_{\pi(i)})\right)^p + (n-m)\cdot c^p\right)\right]^{\frac{1}{p}}, m\leq n \quad (13)$$

$$OSPA_{p,c}(X,\hat{X}) = OSPA_{p,c}(\hat{X},X), m>n$$

Where $\Pi_n$ represents all permutations and combinations of $m$ elements from the set $X$, and the number is $P_n^m(m\leq n)$; $\min_{\pi\in\Pi_n}$ means to find out the combination with the smallest distance difference between all real states and estimated targets; where $p(1\leq p\leq\infty)$ is a distance weight parameter, and $c(c>0)$ is the association weight parameter.

## C. Experiment Results and Analysis

In order to demonstrate the effectiveness of the DeepDA algorithm, it is compared with two classical data association algorithms, JPDA and HA in simulation experiments. At the same time, in order to compare the performance of data association algorithms fairly, the tracking filter consistently uses the same Kalman filter algorithm, and tracks are updated by the weighted output of different data association methods.

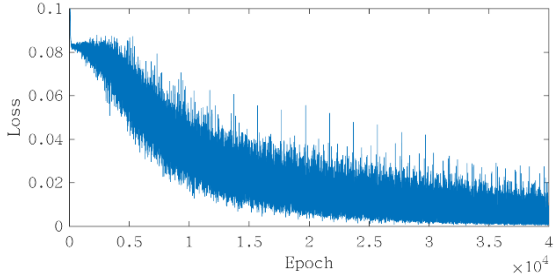

Fig. 8. Loss convergence curve when training

The DeepDA network is trained with an LSTM network with 1 hidden layer. The network is trained on the computer configured with NVIDIA Titan Xp GPU and the whole training takes approximately 5 minutes. The loss function is converged during training shown in Fig. 8.

In the simulated scenario, the results of measurement-track-association algorithms in multi-target tracking using three different data association algorithms are summarized in TABLE II, where parameters of OSPA are set as $c=1$ and $p=2$, respectively.

TABLE II. PERFORMANCE COMPARISON OF DIFFERENT DATA ASSOCIATION ALGORITHMS

| Methods | OSPA | STTI | Time (s) |
|---|---|---|---|
| HA | 0.462 | 1.072 | 0.0335 |
| JPDA | 0.399 | 0.980 | 0.0403 |
| **DeepDA** | **0.394** | **0.844** | **0.0020** |

Two comparative experiments were designed with variable detection probability and variable clutter density. One experiment is setup as the target detection probability is $P_d = 0.9$, the clutter density is set $E_\lambda = 5$; the other experiment is setup as the target detection probability is $P_d = 0.8$, the clutter density is set $E_\lambda = 20$. One of the experimental results of multi-target data association with clutter is shown in Fig. 9. The DeepDA tracker in Fig. 9(d) shows a more robust association on intersecting targets, but the experiments on a single maneuverable target association in clutter indicate their performance is similar.

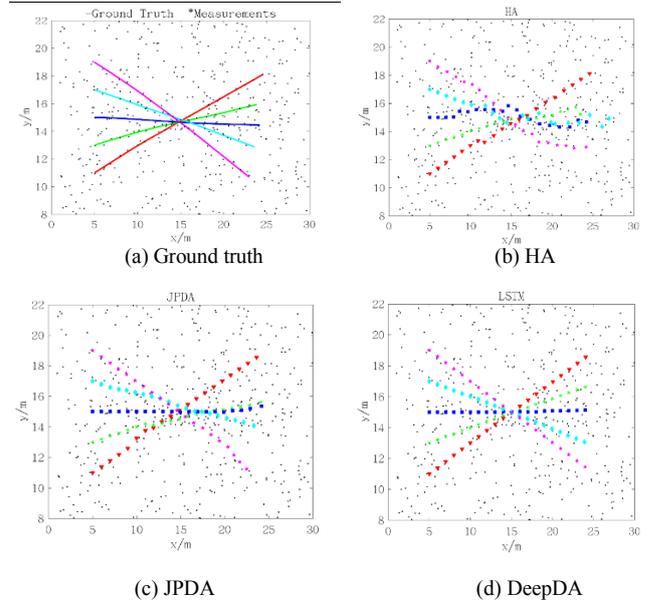

(a) Ground truth (b) HA

(c) JPDA (d) DeepDA

Fig. 9. Multi-target data association results when $P_d = 0.8, E_\lambda = 20$

Fig. 10 and Fig. 11 show the distribution of the OSPA distribution at different clutter density or detection probability.

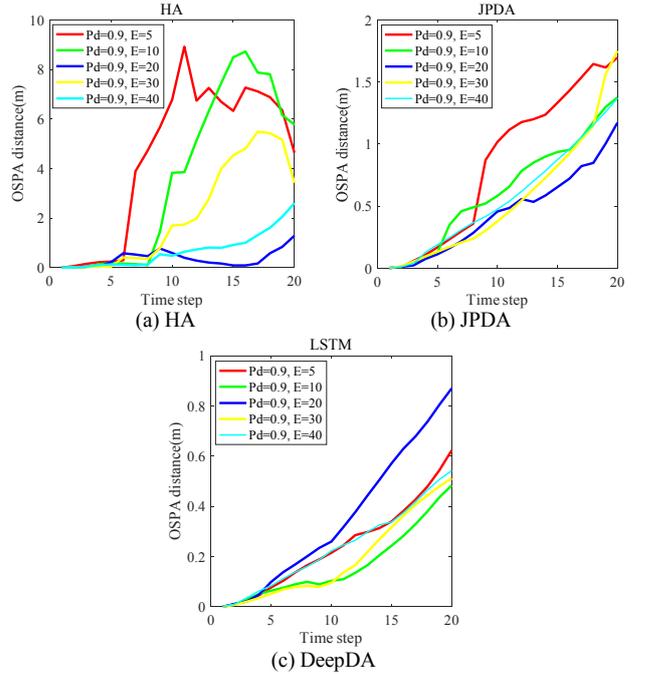

(a) HA (b) JPDA

(c) DeepDA

Fig. 10. The OSPA distance of data association algorithms at different clutter density

Different tracking test samples are designed to test these algorithms. One case is a fixed detection probability, the clutter density is variable, and the other case is just the opposite. And parameters of OSPA are set as $c = 10$ and $p = 2$, respectively.

It can be seen from the scale of y-axis in Fig. 10 that the LSTM algorithm's OSPA is smaller than others, and the variation of clutter density has less influence on the OSPA distance of LSTM at a certain $P_d$. And the HA algorithm is more sensitive to clutter than other methods, but from a single experimental curve, the larger the $E_\lambda$ is, the smaller the OSPA distance is. However, in Fig. 10(c) the OSPA of DeepDA with moderate clutter ($P_d=0.9$ and $E_\lambda=20$) is even higher than cases with higher clutter rate, it can be found that the association probability is also influenced by the spatial distribution of clutter points and tracks in an experiment.

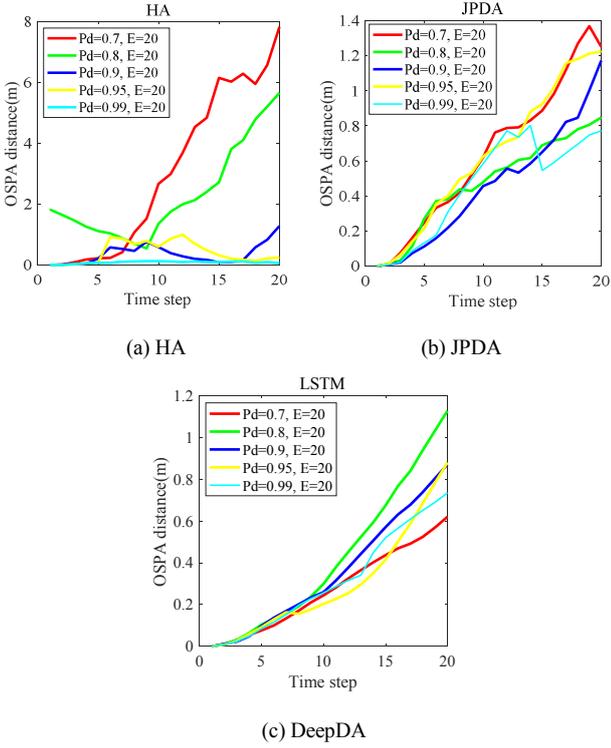

Fig. 11. The OSPA distance of data association algorithms at different detection probability

Similarly, it can be seen from the scale of y-axis in Fig. 11 that the LSTM algorithm's OSPA is smaller than others, and the variation of the target detection probability has less influence on the OSPA distance at a certain $E_\lambda$. Overall, the OSPA distance of the HA algorithm is larger than others, and the higher the detection probability, the smaller the OSPA distance; the OSPA distance of the JPDA algorithm is between the LSTM algorithm and the HA algorithm.

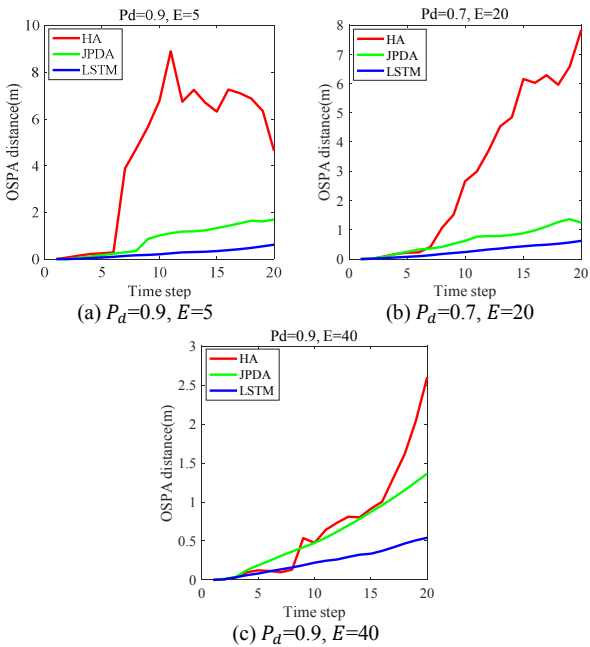

Fig. 12. The OSPA distance of different data association algorithms

Moreover, it can be concluded from Fig. 12 that the LSTM algorithm's OSPA is more stable and generally smaller than the others with different detection probability $P_d$ and clutter density $E_\lambda$. From a single experiment, different $P_d$ and $E_\lambda$ has more influence on the HA algorithm than the others, and it can be found after tracking results analysis of Fig. 12(a) that the HA algorithm is sensitive to random clutter even if the clutter is not too much. It can be concluded from Fig. 10 to Fig. 12 that our proposed algorithm is less affected by the environment than the other two algorithms under the different detection probabilities and clutter densities.

The statistical results of 100 Monte Carlo experiments performed on each group of experiments at different detection probability $P_d$ and clutter density $E_\lambda$ are shown in TABLE III and TABLE IV.

TABLE III. ALGORITHM PERFORMANCE COMPARISON AT DIFFERENT CLUTTER DENSITY ($P_d = 0.9$)

| Metric | Methods | $E_\lambda = 5$ | $E_\lambda = 10$ | $E_\lambda = 20$ | $E_\lambda = 30$ | $E_\lambda = 40$ |
|---|---|---|---|---|---|---|
| OSPA | HA | 1.98±0.72 | 1.45±0.85 | 0.96±0.75 | 0.93±0.80 | 0.55±0.48 |
|  | JPDA | **0.45±0.38** | **0.41±0.61** | 0.42±0.31 | 0.44±0.21 | 0.38±0.26 |
|  | DeepDA | 0.63±0.56 | 0.51±0.44 | **0.37±0.38** | **0.30±0.21** | **0.34±0.25** |
| STTI | HA | 1.10±0.74 | 1.10±0.89 | 1.10±0.84 | 1.10±0.57 | 1.30±0.95 |
|  | JPDA | 1.50±0.71 | 1.59±0.81 | 1.44±0.81 | 1.10±0.57 | 1.70±0.92 |
|  | DeepDA | **1.08±0.92** | **0.97±0.96** | **0.93±0.79** | **1.00±0.94** | **1.20±0.42** |
| Time (s) | HA | 0.0259 | 0.0289 | 0.0339 | 0.0427 | 0.0486 |
|  | JPDA | 0.0148 | 0.0189 | 0.0309 | 0.0396 | 0.0494 |
|  | DeepDA | **0.0108** | **0.0108** | **0.0107** | **0.0107** | **0.0105** |

Some conclusions can be got from TABLE III and TABLE IV. On the STTI metric, compared with classical algorithms like JPDA and HA, our proposed DeepDA algorithm has the lowest STTI under various settings. And on the OSPA metric, our algorithm has superior results under larger clutter density. On the time consuming, HA and JPDA will increase with the increase of clutter density, but our algorithm is basically constant-time consuming under different clutter densities with an average of 0.011 seconds. In summary, the DeepDA algorithm in this paper is more accurate and less time-consuming in multi-target tracking with clutter.

TABLE IV. ALGORITHM PERFORMANCE COMPARISON AT DIFFERENT TARGET DETECTION PROBABILITY ($E_\lambda = 20$)

| Metric | Methods | $P_d = 0.7$ | $P_d = 0.8$ | $P_d = 0.9$ | $P_d = 0.95$ | $P_d = 0.99$ |
|---|---|---|---|---|---|---|
| OSPA | HA | 2.85±1.74 | 2.24±1.08 | 0.96±0.75 | 0.53±0.26 | **0.11±0.04** |
|  | JPDA | **0.50±0.60** | **0.47±0.38** | 0.42±0.31 | 0.37±0.37 | 0.36±0.23 |
|  | DeepDA | 0.63±0.40 | 0.49±0.36 | **0.37±0.38** | **0.36±0.27** | 0.35±0.30 |
| STTI | HA | 1.10±1.10 | 1.30±0.67 | 1.10±0.84 | **1.00±0.94** | 1.50±0.71 |
|  | JPDA | 2.00±0.67 | 1.90±0.74 | 1.44±0.81 | 1.60±0.52 | 1.60±0.84 |
|  | DeepDA | **1.00±0.82** | **0.70±0.82** | **0.93±0.79** | 1.10±0.74 | **0.80±0.63** |

## V. CONCLUSIONS

An LSTM-based deep data association method named as DeepDA in this paper can learn the association probability between multi-target and sensor measurements completely from the measurements with false alarms and missed-detections. Based on an LSTM network, multiple processing layers are constructed in this network to complete the data association between multi-targets and multi-measurements by predicting the association probability. Through a large number of simulation experiments and analysis, it is concluded that the DeepDA algorithm in this paper has following merits: (1) the network can be completely learned from the data series without any prior model, like clutter density, track gating, filter covariance, or other priori information; (2) having more accuracy association results than HA and JPDA algorithm at most situation; (3) less time-consuming than others algorithms, has greater computational efficiency.

ACKNOWLEDGMENT

This research is supported in part by National Natural Science Foundation of China (NSFC) under grant 61402237.